\documentclass{article}
\usepackage{arxiv}
\usepackage{amsfonts} 
\usepackage{arxiv}
\usepackage{amsmath,amssymb,amsthm}
\usepackage[utf8]{inputenc} 
\usepackage[T1]{fontenc} 
\usepackage{microtype}  
\usepackage{caption}
\usepackage{booktabs} 
\usepackage{nicefrac}
\usepackage[utf8]{inputenc}
\usepackage{subcaption}
\usepackage{csquotes}
\usepackage{xcolor}
\usepackage{graphicx}
\usepackage{amsmath}
\usepackage{algorithm}
\usepackage[noend]{algorithmic}
\usepackage{float}
\usepackage{soul}
\usepackage{tikz}
\usepackage{pgfplots}
\AtBeginDocument{%
  \providecommand\BibTeX{{%
    \normalfont B\kern-0.5em{\scshape i\kern-0.25em b}\kern-0.8em\TeX}}}

\sethlcolor{lightgray}
\title{Evolutionary Diversity Optimisation in Constructing  Satisfying Assignments}

\author{Adel Nikfarjam \\
Optimisation and Logistics\\
School of Computer and Mathematical Sciences\\
The University of Adelaide\\
Adelaide, Australia\\
\And
Ralf Rothenberger\\
Chair for Algorithm Engineering\\
Hasso Plattner Institute\\
University of Potsdam\\
Potsdom, Germany\\
\And
Frank Neumann\\
Optimisation and Logistics\\
School of Computer and Mathematical Sciences\\
The University of Adelaide\\
Adelaide, Australia\\
\And
Tobias Friedrich\\
Chair for Algorithm Engineering\\
Hasso Plattner Institute\\
University of Potsdam\\
Potsdom, Germany\\
}

\begin{document}
\maketitle
\begin{abstract}
Computing diverse solutions for a given problem, in particular evolutionary diversity optimisation~(EDO), is a hot research topic in the evolutionary computation community. This paper studies the Boolean satisfiability problem (SAT) in the context of EDO. SAT is of great importance in computer science and differs from the other problems studied in EDO literature, such as KP and TSP. SAT is heavily constrained, and the conventional evolutionary operators are inefficient in generating SAT solutions. Our approach avails of the following characteristics of SAT: 1) the possibility of adding more constraints (clauses) to the problem to forbid solutions or to fix variables, and 2) powerful solvers in the literature, such as minisat. We utilise such a solver to construct a diverse set of solutions.

Moreover, maximising diversity provides us with invaluable information about the solution space of a given SAT problem, such as how large the feasible region is. In this study, we introduce evolutionary algorithms~(EAs) employing a well-known SAT solver to maximise diversity among a set of SAT solutions explicitly. The experimental investigations indicate the introduced algorithms' capability to maximise diversity among the SAT solutions.  
\end{abstract}







\section{Introduction}
Combining the principle of EAs and diversity mechanisms has received increasing attention in the evolutionary computation community. Diversity is widely believed to be essential for survival in dynamic environments. In recent years, the benefits of having access to diverse solutions have been discussed in several studies such as ~\cite{neumann2019evolutionary,NikfarjamB0N21b}. We can categorise the advantage into three main groups: 1) Robustness against dynamic changes and imperfect modeling, 2) critical information about the solution space, and 3) increasing decision-makers' ability to consider and choose between diverse alternatives. 

\subsection{Related Studies}
Traditionally, diversity is seen as a means to avoid premature convergence or explore niches in fitness landscapes of optimisation problems. Niching is a technique that usually divides the population into sub-populations. This enables the algorithms to explore and cover a broader range of solution space. Li et al. \cite{LiEDE17} provides a comprehensive review of niching methods. Two other paradigms, Quality Diversity (QD) and EDO, have recently evolved.

QD aims to compute a diverse set of high-quality solutions that differs in terms of some pre-defined behavioural characteristics. In fact, QD sees diversity in exploring best-performing solutions in a behavioural space. QD has been mostly studied in robotics \cite{RakicevicCK21,ZardiniZZIF21, AllardSCC22}, and game designs \cite{SteckelS21,FontaineTNH20,FontaineLKMTHN21}. Recently, some studies applied QD's principles to combinatorial optimisation problems \cite{NikfarjamMap,Bossek022, NikfarjamDN22}.

EDO is another concept recently developed around the idea of diversity. In contrast to the other paradigms, EDO explicitly maximises the structural diversity of solutions, often subject to a constraint on the solutions' quality. The concept has been defined in \cite{ulrich2011maximizing}, which studied an optimisation problem in continuous domains. Afterwards, EDO has been adapted to generating benchmark instances for TSP and a diverse set of images respecting different aesthetics \cite{alexander2017evolution,doi:10.1162/evcoa00274}. These studies were followed up by works on the use of star-discrepancy, and multi-objective indicators \cite{neumann2018discrepancy,neumann2019evolutionary}. Bossek rt al. \cite{bossek2019evolving} studied the performance of sophisticated mutations at generating a diverse set of benchmark instances for TSP. 

Recently, the focus of the literature shifted from generating instances to computing solutions for combinatorial optimisation problems. There are several problems studied in that regard, such as the traveling salesperson problem (TSP) \cite{viet2020evolving,NikfarjamBN021a}, the knapsack problem (KP)~\cite{BossekN021KP}, the quadratic assignment problem~\cite{DoGNN22}, the minimum spanning tree problem~\cite{Bossek021tree}, the traveling thief problem \cite{NikfarjamTTPEDO}, the optimisation of monotone Sub-modular Functions~\cite{NeumannB021}, and the patient scheduling problem \cite{NikfarjamPAS22}. In most of the mentioned papers, it has been assumed that we already know the optimal solution. The case of unknown optimal solutions has been studied in \cite{NikfarjamB0N21b,NikfarjamCoTTP,neumann2022coevolutionary} by using co-evolutionary techniques or by dividing the population into two subpopulations.  

This paper studies the SAT problem in the context of EDO. SAT is a classical problem in mathematical logic and computer science. The goal of the problem is to determine if there is an assignment to Boolean variables such that a given Boolean formula evaluates to true. The decision variant of SAT is one of the most well-known and well-studied NP-complete problems \cite{Cook71}. One can find many applications for SAT, such as software verification and constraint solving. \cite{DavisLL62} is one of the earliest studies carried out in SAT and introduced a method to compute a satisfying assignment. Several efficient approaches have been developed for SAT in recent years, like Conflict-Driven Clause-Learning (CDCL)~\cite{Silva99grasp} and the Variable State Independent Decaying Sum (VSIDS) branching heuristic~\cite{Moskiewicz01Chaff}. One very versatile solver incorporating those heuristics is minisat~\cite{EenS03}. Minisat has been widely adopted and used as the benchmark solver in the literature. To the best of our knowledge, Nadal \cite{Nadel11} is the only study focusing on the diversity of SAT solutions. They studied \textsc{Diverse}$k$\textsc{Set}, the problem of finding a given number of diverse solutions to an SAT problem, by adapting the variable ordering strategy. However, there can be found another paradigm in the SAT literature, called uniform solution sampling. They aim to compute different solutions without taking diversity into account directly. UniGen2 \cite{ChakrabortyFMSV15} and QuickSampler \cite{DutraLBS18} can be cited here.      

\subsection{Our Contribution}

Several characteristics distinguish SAT from other problems studied in the EDO literature. For instance, the other problems contain either no or few constraints, such as the KP and the TSP. SAT, however, is a highly constrained problem, making it extremely difficult to generate a feasible solution with conventional operators and algorithms in the literature of EDO. In other fields such as constrained programming, researchers often forbid some variables or elements of a given problem to construct a diverse set of solutions. This paper makes a bridge between this approach and EDO. 
Instead of using conventional operators, which are inefficient in SAT, we introduce evolutionary algorithms (EAs) and operators that iteratively modify the original SAT problem by adding clauses. We use a time-efficient solver, such as minisat, to construct new solutions and utilise EDO approaches to maximise the diversity of the solutions. We define two entropy-based diversity measures to quantify the diversity of SAT assignments. The first measure treats all variables equally, while the other takes the frequency of variables in clauses into account. We also conduct a comprehensive experimental investigation, the goal of which is twofold: First, to evaluate the algorithms' performance in constructing diverse assignments. And second, to study the correlation among diversity, solution space, and the number of clauses. For this purpose, we use an SAT generator to construct instances with particular characteristics. Then, we observe how the changes in these characteristics affect the diversity of solutions and algorithms' performances. For example, The introduced mutation outperforms the crossover in the power law SAT instances, while it is the opposite in the uniform instances.       

The remainder of the paper is structured as follows: We first define SAT and diversity in Section \ref{Sec:prob_def}. The diversity algorithms are introduced in Section~\ref{Sec:alg}. The Comprehensive experimental investigation is presented in Section \ref{Sec:EXP}. Finally, we finish with concluding remarks. 

\section{SAT and Diversity}
\label{Sec:prob_def}
The Boolean Satisfiability Problem~(SAT) consists of determining the existence of an assignment (also called model, interpretation, or solution) satisfying a Boolean formula. A Boolean formula is several literals combined by logical connectives, AND ($\land$), and OR ($\lor$), and a literal is a Boolean variable or a negation of a variable ($\neg$). A formula that is formed by the conjunction of a number of clauses (a disjunction of literals) is in conjunctive normal form (CNF). A formula in CNF is satisfiable if there is at least one assignment of the variables such that the formula evaluates to true. In other words, a given CNF formula $\Phi$ is true if an assignment $x$ satisfies all clauses in $\Phi$; otherwise, $\Phi$ is false. This paper aims to compute a diverse set of assignments for a given formula. For this purpose, we require a measure to quantify the diversity of assignments.

\subsection{Diversity}
We utilise an entropy-based measure of diversity. First, we define some notations. Let $X$ denote the set of Boolean variables, $x = (x_1, \cdots , x_n)$ the assignment, and $P$ a set of assignments, where $|X| = n$, $|P| = \mu$, $m$ is number of the clauses. Also, let $f(x_i)$ be the number of assignments in $P$, where $x_i = True$. Then, we can calculate the contribution of each variable to diversity as
\begin{equation*}
h(x_i) = 
\left\{
\begin{array}{ll}
     0 & \mbox{if }f(x_i) = 0\text{ and}\\
     -\left(\frac{f(x_i)}{\mu}\right) \cdot\ln\left(\frac{f(x_i)}{\mu}\right)&
     \mbox{if }f(x_i) > 0\text{.}
\end{array}
\right.
\end{equation*}

In line with EDO literature \cite{NikfarjamBN021a, NikfarjamPAS22}, the entropy of $P$ can be calculated by summation of the variables' contributions:
$$H_1(P) = \sum_{x_i\in X} h(x_i)$$
Nevertheless, some variables appear in clauses more frequently than others. Such variables are likely to be more challenging to diversify, and often play a more important role in the problem. It would be intriguing to give more frequent variables more weight in the entropy calculation such that we first increase such variables' chance to be diverse and second the measure shows the diversity based on the frequency. Therefore, we define the second measure as follows:
$$H_2(P) = \sum_{x_i\in X} r(x_i)\cdot h(x_i),$$
where $r(x_i)$ is the number of occurrences of $x_i$ in the formula. It is beneficial to know the maximum diversity for the measures. It can be used as an upperbound to evaluate a diversity of a set of solutions. We can calculate the optimal $f(x)$ from $\frac{dh(x)}{df(x)} = 0$; Thus, the contribution of a variable is at maximum when:

$$f(x) = \mu \cdot e^{-1}$$

Let denote the optimal $f(x)$ by $f^*$. Since there is no limitations on the number of true variables in $P$ , $H_1$ and $H_2$ are maximum when $\{f(x) = f^* | \forall x \in X\}$. Then, we can calculate $H_1^{max}$ and $H_2^{max}$ form :

$$H_1^{max} = n\cdot f^*$$
$$H_2^{max} = C\cdot f^*$$

where $C$ is the number of the literals in $\Phi$.

\section{Diversity Algorithms}
\label{Sec:alg}
In this paper, we compute a diverse set of assignments for a given SAT problem using the well-known SAT solver minisat. A basic approach to compute $P$ for an SAT problem is to forbid the current assignment by adding a clause to the formula and using the solver to generate another one. For constructing the clause, we can easily make a disjunction of the literals where each literal is the flipped associated variable in the assignment. This method only sometimes leads to a diverse set of assignments. Algorithm \ref{alg:naive} outlines the steps required for this approach.

\begin{algorithm}
\begin{algorithmic}[1]
\WHILE{$|P| < \mu$}
    \STATE Solve the SAT problem by the solver.
    \IF{A satisfying assignment $x$ was found}
    \STATE Add $x$ to $P$.
    \STATE Add a clause forbidding $x$ to $\Phi$.
    \ELSE
    \STATE Break.
    \ENDIF
\ENDWHILE
\end{algorithmic}
\caption{The basic algorithm}
\label{alg:naive}
\end{algorithm}

EDO is another method to compute a diverse set of assignments. We can fix some variables to true or false and then use the solver (minisat) to determine a satisfying assignment with those fixed variables. Afterwards, we can employ EDO approaches to maximise diversity. Here, the question is how to choose the fixed variables. In line with most EDO algorithms in the literature, we can randomly select one of the current solutions and, by standard bit flip mutation, flip some of the variable assignments and fix them. In contrast to the standard bit-flip mutation, where the rest of the variables remain unchanged, the solver determines the value for the other variables.
\begin{algorithm}
\begin{algorithmic}[1]
\STATE Solve the SAT problem by the solver, and add $x$ to $P$.
\WHILE{A termination criterion is met}
    \STATE Select an assignment $x$ from $P$ uniformly at random.
    \STATE Select and flip  each variable independently with probability $\frac{1}{n}$.
    \STATE Add clauses fixing the selected variables to $\Phi$
    \STATE Solve $\Phi$ and determine unfixed variables by the solver.
    \IF{A satisfying assignment $x$ was found}
    \IF{$|P| > \mu$}
    \STATE Add $x$ to $P$.
    \STATE Remove one individual $x$ from $P$, where $x = \arg \max_{x \in P} H(P \setminus \{x\})$.
    \ENDIF
    \ENDIF
    \STATE Remove the clauses that fixing the variables from $\Phi$.
\ENDWHILE
\end{algorithmic}
\caption{The bit-flip evolutionary algorithm}
\label{alg:basic}
\end{algorithm}
Algorithm \ref{alg:basic} describes this approach. First, we find the first satisfying assignment for $\Phi$ by minisat and add it to $P$. Then, we select a solution in $P$ uniformly at random and choose and flip some variables by the bit-flip mutation. After adding clauses to $\Phi$ that fix the selected variables, we solve $\Phi$ by minisat. If a satisfying assignment is found, we add it to $P$; Then, if $|P| > \mu$, we remove an assignment $x$ with the least contribution to the diversity of $P$. Finally, we remove the clauses fixing the variables from $\Phi$. We repeat these steps until a termination criterion is met.    

\begin{algorithm}
\begin{algorithmic}[1]
\WHILE{$|P| < \mu$}
    \STATE Randomly fix $l$ variables (determine $y$).
    \STATE Add the clauses that fix the variables in $y$ to $\Phi$ and solve it by the solver.
    \IF{A satisfying assignment $x$ was found}
    \STATE Add $x$ to $P$ and $y$ to $Y$.
    \ENDIF
    \STATE Remove the clauses fixing the variables from $\Phi$.
\ENDWHILE
\WHILE{A termination criterion is met}
    \STATE Randomly select one (two) parent(s) $y_i$ ($y_j$) from $Y$.
    \STATE Generate a new solution $y_o$ by mutation or crossover + mutation. 
    \STATE Add clauses that fix the variables in $y_o$ to $\Phi$ and solve the SAT problem.
    \IF{A satisfying assignment $x$ is found}
    \STATE Add $x$ to $P$ and $y_o$ to $Y$.
    \STATE Remove one individual $x$ from $P$, where $x = \arg \max_{x \in P} H(P \setminus \{x\})$, and the corresponding solution $y$ from $Y$.
    \ENDIF
    \STATE Remove the clauses fixing the variables from $\Phi$.
\ENDWHILE
\end{algorithmic}
\caption{The EDO algorithm}
\label{alg:EDO}
\end{algorithm}
\begin{figure}
\centering
\begin{subfigure}{\linewidth}
\centering
\includegraphics[width=1\columnwidth]{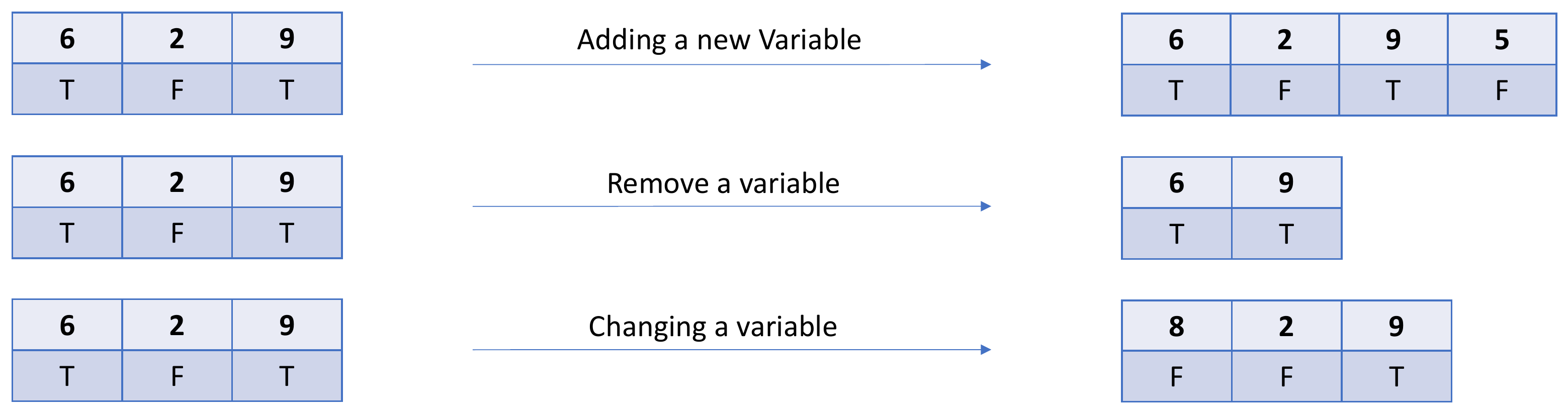}
\caption{Mutation}\vspace{-10pt}
\label{fig:mut}
\end{subfigure}
\begin{subfigure}{\linewidth}
\centering
\vspace{+10pt}
\includegraphics[width=1\columnwidth]{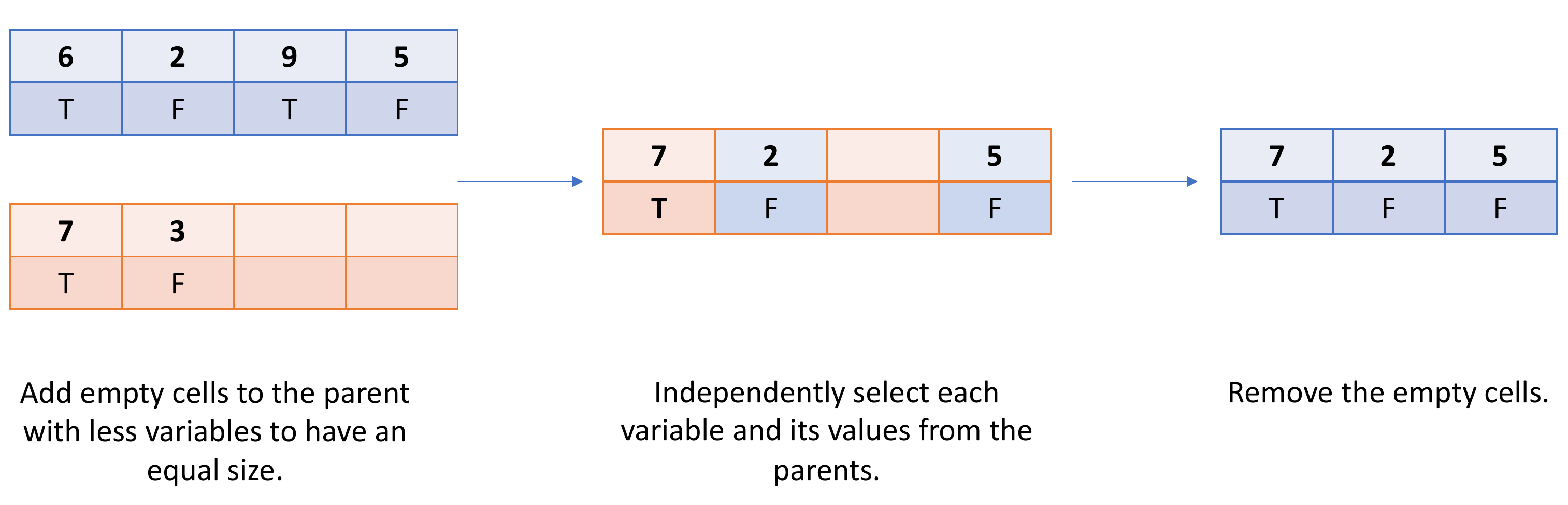}
\caption{Crossover}\vspace{-5pt}
\label{fig:cross}
\end{subfigure}
\caption{The representation of solution $y$, the mutation, and the crossover in the EDO algorithm \ref{alg:EDO}.}\vspace{-10pt}
\label{fig:represent}
\end{figure}
Since minisat is an exact algorithm, we can map from the fixed variables to the actual assignments. Thus, we can save the fixed variables and operate ~(crossover, mutation) on them. So, we have a solution $y$ consisting of a string $y' = (y'_1, \cdots , y'_l)$ showing the index of fixed variables and a Boolean string $y''$ showing their values. Let $Y$ be a set of solutions $y$, where $|Y| = \mu$. Note that from each $y_i \in Y$ we can map to $x_i \in P$, by fixing variables in $y_i$ and solving the problem by the solver. 

Algorithm~\ref{alg:EDO} sketches the steps required in this approach. The algorithm consists of two stages, the initialisation and the evolutionary stage. In initialisation, we randomly generate a variable $y$, where $|y| = l$. We solve $\phi$ after adding clauses to it. If a satisfying assignment $x$ is found, we add $x$ to $P$, and $y$ to $Y$. Afterwards, we remove the clauses fixing the variables from $\Phi$. we continue these steps until $|P| = \mu$.

Having constructed an initial population, we move to the evolutionary stage. We first select a solution $y$ (or two solutions in case of crossover) from $Y$ and generate an offspring $y_o$ by mutation (or first crossover, then mutation). After adding clauses fixing variables in $y_o$ to $\Phi$, we solve it by the solver. If a satisfying assignment $x$ is found, we add $x$ to $P$ and $y$ to $Y$; then remove a $x$ from $P$ and the corresponding $y$ from $Y$ that has the least contribution to the diversity of $P$. In last step, we remove the clauses fixing the variables from $\Phi$. We repeat these steps in the evolutionary stage until a termination criterion is met.

We now describe the operators, the mutation and the crossover. For the mutation, we take one of the following three actions uniformly at random: 1) Fix another variable (add a new variable to $y$), 2) unfix a variable (remove a variable from $y$), or 3) switch a fixed variable with an unfixed variable, all uniformly at random. The steps are depicted in Figure\ref{fig:mut}. Turning to the crossover, we add empty cells to the parent with fewer fixed variables to make the sizes equal. Then, we select each variable randomly from the parents with probability $1/2$. Figure\ref{fig:cross} illustrates the steps required by the crossover.

\section{Experimental Investigation}
\label{Sec:EXP}
\begin{table*}
\centering
\caption{The diversity obtained from the algorithms using $H_1$ as the fitness function in 30 independent runs. Stat shows the results of Kruskal-Wallis statistical test at a $5\%$ significance level with Bonferroni correction. In row Stat, the notation $X^+$ means the median of the measure ($H_1$) is better than the one for variant $X$, $X^-$ means it is worse, and $X^*$ indicates no significant difference.}

\renewcommand{\tabcolsep}{5.1pt}
\renewcommand{\arraystretch}{1.1}
\begin{tabular}{l|ccc|ccc|ccc|ccc}
\toprule
        &\multicolumn{3}{c|}{Basic \ref{alg:naive}}&\multicolumn{3}{c|}{Bit-flip \ref{alg:basic}}&\multicolumn{3}{c|}{EDO \ref{alg:EDO} \begin{scriptsize}Mutation\end{scriptsize}}&\multicolumn{3}{c}{EDO \ref{alg:EDO} \begin{scriptsize}
            Crossover+Mutation
        \end{scriptsize}}\\
\cmidrule(l{2pt}r{2pt}){1-1} 
\cmidrule(l{2pt}r{2pt}){2-4}
\cmidrule(l{2pt}r{2pt}){5-7}
\cmidrule(l{2pt}r{2pt}){8-10}
\cmidrule(l{2pt}r{2pt}){11-13}

             m & $H_1$ & $H_2$ & Stat (1) & $H_1$ & $H_2$ & Stat (2) & $H_1$ & $H_2$ & Stat (3) & $H_1$ & $H_2$ & Stat (4) \\
\midrule
210&0.055&0.016&$2^-3^-4^-$&0.753&0.839&$1^+3^-4^-$&\hl{0.962}&0.959&$1^+2^+4^*$&0.953&0.955&$1^+2^+3^*$\\
220&0.052&0.011&$2^-3^-4^-$&0.721&0.818&$1^+3^-4^-$&\hl{0.945}&0.938&$1^+2^+4^*$&0.932&0.933&$1^+2^+3^*$\\
230&0.055&0.019&$2^-3^-4^-$&0.738&0.823&$1^+3^-4^-$&\hl{0.937}&0.932&$1^+2^+4^*$&0.925&0.925&$1^+2^+3^*$\\
240&0.046&0.007&$2^-3^-4^-$&0.731&0.808&$1^+3^-4^-$&\hl{0.933}&0.927&$1^+2^+4^*$&0.921&0.924&$1^+2^+3^*$\\
250&0.171&0.135&$2^-3^-4^-$&0.774&0.851&$1^+3^-4^-$&\hl{0.928}&0.918&$1^+2^+4^*$&0.911&0.915&$1^+2^+3^*$\\
260&0.114&0.075&$2^-3^-4^-$&0.765&0.832&$1^+3^-4^-$&\hl{0.925}&0.909&$1^+2^+4^*$&0.914&0.904&$1^+2^+3^*$\\
270&0.089&0.061&$2^-3^-4^-$&0.757&0.823&$1^+3^-4^-$&\hl{0.911}&0.893&$1^+2^+4^*$&0.896&0.886&$1^+2^+3^*$\\
280&0.172&0.143&$2^-3^-4^-$&0.76&0.828&$1^+3^-4^-$&\hl{0.907}&0.897&$1^+2^+4^*$&0.886&0.885&$1^+2^+3^*$\\
290&0.14&0.083&$2^-3^-4^-$&0.826&0.842&$1^+3^-4^-$&\hl{0.912}&0.878&$1^+2^+4^+$&0.9&0.874&$1^+2^+3^-$\\
300&0.272&0.235&$2^-3^-4^-$&0.825&0.825&$1^+3^-4^-$&\hl{0.902}&0.856&$1^+2^+4^*$&0.895&0.857&$1^+2^+3^*$\\
310&0.191&0.156&$2^-3^-4^-$&0.776&0.777&$1^+3^-4^*$&\hl{0.862}&0.814&$1^+2^+4^*$&0.844&0.806&$1^+2^*3^*$\\
320&0.099&0.051&$2^-3^-4^-$&0.478&0.424&$1^+3^-4^*$&\hl{0.611}&0.489&$1^+2^+4^*$&0.591&0.478&$1^+2^*3^*$\\
330&0.169&0.135&$2^-3^-4^-$&0.544&0.503&$1^+3^-4^*$&\hl{0.666}&0.56&$1^+2^+4^*$&0.643&0.547&$1^+2^*3^*$\\
340&0.182&0.129&$2^-3^-4^-$&0.627&0.562&$1^+3^-4^-$&\hl{0.73}&0.611&$1^+2^+4^*$&0.717&0.603&$1^+2^+3^*$\\
350&0.157&0.113&$2^-3^-4^-$&0.534&0.496&$1^+3^-4^*$&\hl{0.61}&0.532&$1^+2^+4^*$&0.605&0.531&$1^+2^*3^*$\\
360&0.089&0.047&$2^-3^-4^-$&0.531&0.501&$1^+3^-4^*$&\hl{0.606}&0.537&$1^+2^+4^*$&0.6&0.535&$1^+2^*3^*$\\
370&0.156&0.11&$2^-3^-4^-$&0.425&0.339&$1^+3^-4^-$&\hl{0.535}&0.394&$1^+2^+4^*$&0.529&0.392&$1^+2^+3^*$\\
380&0.161&0.121&$2^-3^-4^-$&0.437&0.344&$1^+3^-4^*$&\hl{0.498}&0.375&$1^+2^+4^*$&0.491&0.372&$1^+2^*3^*$\\

\bottomrule
\end{tabular}
\label{tbl:Res_powerlaw}
\end{table*}
\begin{table*}
\centering
\caption{The diversity obtained from the algorithms using $H_1$ as the fitness function. The variables appear in clauses based on a uniform distribution with $n = 100$ and $k = 3$. The notations are in line with Table \ref{tbl:Res_powerlaw}}

\renewcommand{\tabcolsep}{5.1pt}
\renewcommand{\arraystretch}{1.1}
\begin{tabular}{l|ccc|ccc|ccc|ccc}
\toprule
        &\multicolumn{3}{c|}{Basic \ref{alg:naive}}&\multicolumn{3}{c|}{Bit-flip \ref{alg:basic}}&\multicolumn{3}{c|}{EDO \ref{alg:EDO} \begin{scriptsize}Mutation\end{scriptsize}}&\multicolumn{3}{c}{EDO \ref{alg:EDO} \begin{scriptsize}
            Crossover+Mutation
        \end{scriptsize}}\\
\cmidrule(l{2pt}r{2pt}){1-1} 
\cmidrule(l{2pt}r{2pt}){2-4}
\cmidrule(l{2pt}r{2pt}){5-7}
\cmidrule(l{2pt}r{2pt}){8-10}
\cmidrule(l{2pt}r{2pt}){11-13}

             m & $H_1$ & $H_2$ & Stat (1) & $H_1$ & $H_2$ & Stat (2) & $H_1$ & $H_2$ & Stat (3) & $H_1$ & $H_2$ & Stat (4) \\
\midrule

270&0.295&0.28&$2^-3^-4^-$&0.859&0.889&$1^+3^-4^-$&\hl{0.942}&0.947&$1^+2^+4^*$&0.94&0.948&$1^+2^+3^*$\\
280&0.241&0.217&$2^-3^-4^-$&0.867&0.879&$1^+3^-4^-$&0.944&0.943&$1^+2^+4^*$&0.944&0.946&$1^+2^+3^*$\\
290&0.202&0.186&$2^-3^-4^-$&0.834&0.848&$1^+3^-4^-$&0.937&0.938&$1^+2^+4^*$&\hl{0.939}&0.941&$1^+2^+3^*$\\
300&0.183&0.175&$2^-3^-4^-$&0.877&0.888&$1^+3^-4^-$&0.943&0.943&$1^+2^+4^*$&\hl{0.946}&0.946&$1^+2^+3^*$\\
310&0.09&0.078&$2^-3^-4^-$&0.875&0.893&$1^+3^-4^-$&0.943&0.946&$1^+2^+4^*$&\hl{0.945}&0.948&$1^+2^+3^*$\\
320&0.062&0.051&$2^-3^-4^-$&0.884&0.894&$1^+3^-4^-$&0.936&0.939&$1^+2^+4^*$&\hl{0.937}&0.94&$1^+2^+3^*$\\
330&0.157&0.137&$2^-3^-4^-$&0.885&0.895&$1^+3^-4^-$&0.927&0.927&$1^+2^+4^*$&\hl{0.932}&0.934&$1^+2^+3^*$\\
340&0.135&0.117&$2^-3^-4^-$&0.898&0.905&$1^+3^-4^-$&0.928&0.927&$1^+2^+4^*$&\hl{0.933}&0.933&$1^+2^+3^*$\\
350&0.073&0.062&$2^-3^-4^-$&0.895&0.903&$1^+3^-4^-$&0.916&0.918&$1^+2^+4^*$&\hl{0.918}&0.92&$1^+2^+3^*$\\
360&0.08&0.067&$2^-3^-4^-$&0.866&0.875&$1^+3^-4^-$&0.893&0.896&$1^+2^+4^*$&\hl{0.898}&0.903&$1^+2^+3^*$\\
370&0.084&0.07&$2^-3^-4^-$&0.851&0.862&$1^+3^-4^-$&0.884&0.886&$1^+2^+4^*$&\hl{0.891}&0.895&$1^+2^+3^*$\\
380&0.058&0.042&$2^-3^-4^-$&0.846&0.855&$1^+3^-4^-$&0.876&0.879&$1^+2^+4^*$&\hl{0.877}&0.88&$1^+2^+3^*$\\
390&0.178&0.178&$2^-3^-4^-$&0.822&0.822&$1^+3^*4^-$&0.832&0.829&$1^+2^*4^*$&\hl{0.835}&0.832&$1^+2^+3^*$\\
400&0.226&0.215&$2^-3^-4^-$&0.637&0.622&$1^+3^-4^-$&\hl{0.648}&0.63&$1^+2^+4^*$&0.647&0.629&$1^+2^+3^*$\\
410&0.105&0.098&$2^-3^-4^-$&0.674&0.669&$1^+3^-4^-$&0.693&0.685&$1^+2^+4^*$&0.693&0.684&$1^+2^+3^*$\\
420&0.125&0.118&$2^-3^-4^-$&0.603&0.592&$1^+3^*4^-$&0.612&0.599&$1^+2^*4^*$&\hl{0.613}&0.6&$1^+2^+3^*$\\
430&0.153&0.146&$2^-3^-4^-$&0.311&0.299&$1^+3^*4^-$&0.326&0.309&$1^+2^*4^*$&0.326&0.309&$1^+2^+3^*$\\
440&0.059&0.047&$2^-3^-4^-$&0.352&0.335&$1^+3^-4^-$&0.366&0.346&$1^+2^+4^*$&0.366&0.347&$1^+2^+3^*$\\

\bottomrule
\end{tabular}
\label{tbl:Res_uniform}
\end{table*}

This section empirically studies and compares the introduced algorithms. We examine two variations of Algorithm~\ref{alg:EDO}: One solely employs mutation as the operator, while the other first generates an offspring by crossover and then uses mutation on the offspring. To examine the algorithms, we use the SAT generator \cite{AnsoteguiBL09} to generate two sets of CNF formulas. The SAT generator is also used for experimental investigations in \cite{0001KRS17, esa/0001KRSS17}. In the first set, the variables appear in clauses based on a power law distribution. The following parameters were used in generating the first set: $n = 100$, $k = 3$, $\beta = 2.75$, and $m = \{210, 220, \cdots, 380\}$, where $k$ and $\beta$ are the number of literals in a clause and the power law exponent, respectively. In the second set, the variables appear in the clauses based on the uniform distribution. The parameters for the set are: $n = 100$, $k = 3$, and $m = \{270, 280, \cdots, 440\}$. We set $\mu = 20$ and consider $2000$ iterations as the termination criterion for the EAs. Instead of 30 independent runs on one formula, we generate 30 formulas for each configuration and run the algorithms once on each formula. This helps us to comprehend more about SAT instances having the same characteristics. In algorithm \ref{alg:EDO}, $l$ should be set with taking number of clauses into account; a higher number of clauses, a lower $l$. Based on preliminary investigation, We set $l = 10$ for the lowest number of clauses in each set and gradually decrease it to $4$ for the greatest number of clauses. Note that we made sure all formulas were satisfiable ($\Phi = true$).    

\subsection{Comparison of algorithms employing $H_1$ as the fitness}

In this section, we compare the diversity of SAT assignments obtained by the presented algorithms using $H_1$ as the fitness function. Table~\ref{tbl:Res_powerlaw} summarises the algorithms' results in the first set of instances (formulas). As expected, the basic algorithm results in assignments with poor diversity; the $H_1$ values range between $1.71$ and $10.01$. If we normalise these values, the range is from $5\%$ to $27\%$. The interesting information is that the increase in the clause-variable ratio $\frac{m}{n}$ has no meaningful impact on the basic algorithm's result. The expectation is that an increase in $\left(\frac{m}{n}\right)$ reduces the feasible region which leads to a decrease in the diversity of assignments; we can observe the trend in the results of the other algorithms. 

As Table~\ref{tbl:Res_powerlaw} shows, the bit-flip brings about considerably more diverse assignments than the basic algorithm. The observation can be confirmed by the Kruskal-Wallis statistical test at a $5\%$ significance level and with Bonferroni correction. The mean of diversity ranges from $44\%$ to $82\%$. Although there are also fluctuations in the bit-flip algorithm's results, we can observe a general decrease in diversity by an increase in $\left(\frac{m}{n}\right)$, especially when $m$ is larger than $290$. However, if we only consider the first half of the table, it is exactly the other way around; there is a slight increase in diversity obtained. One plausible reason is that the minisat solver is an exact algorithm, and bit-flip mutation does not impose as significant changes as required. On the other hand, an increase in $\left(\frac{m}{n}\right)$ makes even minor changes significantly impact the assignments. In fact, the feasible regain and the maximum achievable diversity decrease in instances with medium values of $\left(\frac{m}{n}\right)$ compared to small ones, but the bit-flip algorithm performs better in these instances.

Table~\ref{tbl:Res_powerlaw} indicates the superiority of EDO algorithms in constructing diverse sets of SAT assignments. Both algorithm variants yield decent results and statistically outperform the basic and the bit-flip algorithms in all instances. Here, we can observe a more static downward trend in diversity with increasing $\frac{m}{n}$. It results in sets with more than $90\%$ diversity (normalised $H_1$) for instances with $\frac{m}{n} \leq 3$. For example, the mean of diversity is $96\%$ in cases where $m = 210$. Interestingly, the variant using only mutation results in slightly higher diversity. Although, it is not statistically significant.

Table~\ref{tbl:Res_uniform} draws a similar comparison between the algorithms on the set of uniform formulas. Almost all our observations in Table~\ref{tbl:Res_powerlaw} are still valid.
Table~\ref{tbl:Res_uniform} shows that: 1) Algorithm~\ref{alg:naive} results in solutions with poor diversity ranging from $6\%$ to $29\%$. Nevertheless, the diversity obtained in the uniform instances is higher compared to the power law formulas. 2) Bit-flip performs better than the basic algorithm but worse than the EDO variants. The average $H_1$ obtained by the bit-flip algorithms ranges from $0.31$ to $0.86$. 3) We can observe a descending trend in diversity for increasing $\frac{m}{n}$, especially in the EDO algorithms' results.

The most interesting part of the table is comparing the two EDO variants. In contrast to the power law instances, the variant using both crossover and mutation slightly outperforms the other one in terms of  $H_1$. We can get diverse sets of SAT assignments with more than $90\%$ diversity in terms of $H_1$ with the EDO algorithm in cases $m \leq 360$.

\subsection{Comparison of algorithms employing $H_2$ as the fitness}

\begin{table*}
\centering
\caption{The diversity obtained from the algorithms using $H_2$ as the fitness function on the same instances in Table \ref{tbl:Res_powerlaw}. The Kruskal-Wallis statistical test is conducted on $H_2$. The notations are in line with Table \ref{tbl:Res_powerlaw}}

\renewcommand{\tabcolsep}{5.1pt}
\renewcommand{\arraystretch}{1.1}
\begin{tabular}{l|ccc|ccc|ccc|ccc}
\toprule
        &\multicolumn{3}{c|}{Basic \ref{alg:naive}}&\multicolumn{3}{c|}{Bit-flip \ref{alg:basic}}&\multicolumn{3}{c|}{EDO \ref{alg:EDO} \begin{scriptsize}Mutation\end{scriptsize}}&\multicolumn{3}{c}{EDO \ref{alg:EDO} \begin{scriptsize}
            Crossover+Mutation
        \end{scriptsize}}\\
\cmidrule(l{2pt}r{2pt}){1-1} 
\cmidrule(l{2pt}r{2pt}){2-4}
\cmidrule(l{2pt}r{2pt}){5-7}
\cmidrule(l{2pt}r{2pt}){8-10}
\cmidrule(l{2pt}r{2pt}){11-13}

             m & $H_2$ & $H_1$ & Stat (1) & $H_2$ & $H_1$ & Stat (2) & $H_2$ & $H_1$ & Stat (3) & $H_2$ & $H_1$ & Stat (4) \\
\midrule

210&0.016&0.055&$2^-3^-4^-$&0.849&0.732&$1^+3^-4^-$&\hl{0.965}&0.944&$1^+2^+4^*$&0.957&0.912&$1^+2^+3^*$\\
220&0.011&0.052&$2^-3^-4^-$&0.83&0.709&$1^+3^-4^-$&\hl{0.95}&0.928&$1^+2^+4^*$&0.939&0.892&$1^+2^+3^*$\\
230&0.019&0.055&$2^-3^-4^-$&0.836&0.719&$1^+3^-4^-$&\hl{0.941}&0.911&\hl{$1^+2^+4^+$}&0.932&0.882&$1^+2^+3^-$\\
240&0.007&0.046&$2^-3^-4^-$&0.825&0.706&$1^+3^-4^-$&\hl{0.936}&0.896&$1^+2^+4^*$&0.929&0.874&$1^+2^+3^*$\\
250&0.135&0.171&$2^-3^-4^-$&0.859&0.757&$1^+3^-4^-$&\hl{0.927}&0.909&$1^+2^+4^*$&0.918&0.874&$1^+2^+3^*$\\
260&0.075&0.114&$2^-3^-4^-$&0.845&0.751&$1^+3^-4^-$&\hl{0.919}&0.906&$1^+2^+4^*$&0.911&0.872&$1^+2^+3^*$\\
270&0.061&0.089&$2^-3^-4^-$&0.838&0.737&$1^+3^-4^-$&\hl{0.907}&0.89&\hl{$1^+2^+4^+$}&0.895&0.844&$1^+2^+3^-$\\
280&0.143&0.172&$2^-3^-4^-$&0.842&0.74&$1^+3^-4^-$&\hl{0.906}&0.877&$1^+2^+4^*$&0.895&0.84&$1^+2^+3^*$\\
290&0.083&0.14&$2^-3^-4^-$&0.861&0.807&$1^+3^-4^-$&\hl{0.895}&0.887&\hl{$1^+2^+4^+$}&0.887&0.864&$1^+2^+3^-$\\
300&0.235&0.272&$2^-3^-4^-$&0.835&0.813&$1^+3^-4^-$&0.865&0.884&$1^+2^+4^*$&0.865&0.867&$1^+2^+3^*$\\
310&0.156&0.191&$2^-3^-4^-$&0.786&0.762&$1^+3^-4^*$&\hl{0.824}&0.845&$1^+2^+4^*$&0.816&0.815&$1^+2^*3^*$\\
320&0.051&0.099&$2^-3^-4^-$&0.434&0.468&$1^+3^-4^*$&\hl{0.494}&0.599&$1^+2^+4^*$&0.481&0.558&$1^+2^*3^*$\\
330&0.135&0.169&$2^-3^-4^-$&0.513&0.534&$1^+3^-4^*$&\hl{0.562}&0.647&$1^+2^+4^*$&0.551&0.61&$1^+2^*3^*$\\
340&0.129&0.182&$2^-3^-4^-$&0.57&0.617&$1^+3^-4^*$&\hl{0.616}&0.718&$1^+2^+4^*$&0.606&0.69&$1^+2^*3^*$\\
350&0.113&0.157&$2^-3^-4^-$&0.505&0.524&$1^+3^-4^*$&\hl{0.537}&0.598&$1^+2^+4^*$&0.534&0.587&$1^+2^*3^*$\\
360&0.047&0.089&$2^-3^-4^-$&0.504&0.524&$1^+3^-4^*$&\hl{0.541}&0.602&$1^+2^+4^*$&0.536&0.589&$1^+2^*3^*$\\
370&0.11&0.156&$2^-3^-4^-$&0.342&0.42&$1^+3^-4^-$&\hl{0.396}&0.53&$1^+2^+4^*$&0.392&0.521&$1^+2^+3^*$\\
380&0.121&0.161&$2^-3^-4^-$&0.347&0.432&$1^+3^-4^*$&\hl{0.377}
&0.494&$1^+2^+4^*$&0.374&0.484&$1^+2^*3^*$\\

\bottomrule
\end{tabular}
\label{tbl:Res_powerlaw_ne}
\end{table*}

\begin{table*}
\centering
\caption{The diversity obtained from the algorithms using $H_2$ the fitness on the same instances in Table \ref{tbl:Res_uniform}. The notations are in line with Table \ref{tbl:Res_powerlaw_ne}.}

\renewcommand{\tabcolsep}{5.1pt}
\renewcommand{\arraystretch}{1.1}
\begin{tabular}{l|ccc|ccc|ccc|ccc}
\toprule
        &\multicolumn{3}{c|}{Basic \ref{alg:naive}}&\multicolumn{3}{c|}{Bit-flip \ref{alg:basic}}&\multicolumn{3}{c|}{EDO \ref{alg:EDO} \begin{scriptsize}Mutation\end{scriptsize}}&\multicolumn{3}{c}{EDO \ref{alg:EDO} \begin{scriptsize}
            Crossover+Mutation
        \end{scriptsize}}\\
\cmidrule(l{2pt}r{2pt}){1-1} 
\cmidrule(l{2pt}r{2pt}){2-4}
\cmidrule(l{2pt}r{2pt}){5-7}
\cmidrule(l{2pt}r{2pt}){8-10}
\cmidrule(l{2pt}r{2pt}){11-13}

             m & $H_2$ & $H_1$ & Stat (1) & $H_2$ & $H_1$ & Stat (2) & $H_2$ & $H_1$ & Stat (3) & $H_2$ & $H_1$ & Stat (4) \\
\midrule

270&0.28&0.295&$2^-3^-4^-$&0.891&0.849&$1^+3^-4^-$&\hl{0.95}&0.933&$1^+2^+4^*$&0.946&0.923&$1^+2^+3^*$\\
280&0.217&0.241&$2^-3^-4^-$&0.881&0.863&$1^+3^-4^-$&0.946&0.937&$1^+2^+4^*$&\hl{0.947}&0.934&$1^+2^+3^*$\\
290&0.186&0.202&$2^-3^-4^-$&0.849&0.83&$1^+3^-4^-$&0.939&0.931&$1^+2^+4^*$&\hl{0.942}&0.93&$1^+2^+3^*$\\
300&0.175&0.183&$2^-3^-4^-$&0.891&0.871&$1^+3^-4^-$&0.945&0.939&$1^+2^+4^*$&\hl{0.947}&0.936&$1^+2^+3^*$\\
310&0.078&0.09&$2^-3^-4^-$&0.897&0.874&$1^+3^-4^-$&0.945&0.933&$1^+2^+4^*$&\hl{0.95}&0.937&$1^+2^+3^*$\\
320&0.051&0.062&$2^-3^-4^-$&0.895&0.881&$1^+3^-4^-$&0.938&0.929&$1^+2^+4^*$&\hl{0.94}&0.928&$1^+2^+3^*$\\
330&0.136&0.157&$2^-3^-4^-$&0.897&0.882&$1^+3^-4^-$&0.93&0.923&$1^+2^+4^*$&\hl{0.934}&0.923&$1^+2^+3^*$\\
340&0.117&0.135&$2^-3^-4^-$&0.907&0.895&$1^+3^-4^-$&0.932&0.927&$1^+2^+4^*$&\hl{0.934}&0.925&$1^+2^+3^*$\\
350&0.062&0.073&$2^-3^-4^-$&0.904&0.892&$1^+3^-4^-$&0.919&0.911&$1^+2^+4^*$&\hl{0.922}&0.913&$1^+2^+3^*$\\
360&0.067&0.08&$2^-3^-4^-$&0.879&0.865&$1^+3^-4^-$&0.897&0.889&$1^+2^+4^*$&\hl{0.904}&0.894&$1^+2^+3^*$\\
370&0.07&0.084&$2^-3^-4^-$&0.866&0.851&$1^+3^-4^-$&0.892&0.882&$1^+2^+4^*$&\hl{0.896}&0.884&$1^+2^+3^*$\\
380&0.042&0.058&$2^-3^-4^-$&0.858&0.843&$1^+3^-4^-$&0.878&0.867&$1^+2^+4^*$&\hl{0.881}&0.869&$1^+2^+3^*$\\
390&0.178&0.178&$2^-3^-4^-$&0.824&0.818&$1^+3^*4^-$&0.832&0.829&$1^+2^*4^*$&\hl{0.834}&0.83&$1^+2^+3^*$\\
400&0.214&0.226&$2^-3^-4^-$&0.623&0.636&$1^+3^-4^-$&0.631&0.646&$1^+2^+4^*$&0.631&0.645&$1^+2^+3^*$\\
410&0.098&0.105&$2^-3^-4^-$&0.672&0.673&$1^+3^-4^-$&0.684&0.688&$1^+2^+4^*$&0.684&0.686&$1^+2^+3^*$\\
420&0.118&0.125&$2^-3^-4^-$&0.593&0.601&$1^+3^-4^*$&\hl{0.602}&0.611&$1^+2^+4^*$&0.601&0.61&$1^+2^*3^*$\\
430&0.146&0.153&$2^-3^-4^-$&0.3&0.311&$1^+3^*4^-$&0.309&0.325&$1^+2^*4^*$&0.309&0.326&$1^+2^+3^*$\\
440&0.047&0.059&$2^-3^-4^-$&0.336&0.352&$1^+3^-4^-$&0.347&0.366&$1^+2^+4^*$&0.347&0.366&$1^+2^+3^*$\\

\bottomrule
\end{tabular}
\label{tbl:Res_uniform_ne}
\end{table*}
We examine the algorithms' performance when $H_2$ is incorporated as the fitness function. The $H_2$ differs from $H_1$ in focusing on the variables with more appearances in $\Phi$. Table \ref{tbl:Res_powerlaw_ne} and \ref{tbl:Res_uniform_ne} summarise the algorithms' results in the power law and uniform instances, respectively. Since Algorithm \ref{alg:basic} does not use any diversity measures inside of the algorithm, the results are the same as those of Table \ref{tbl:Res_powerlaw} and \ref{tbl:Res_uniform}. Nevertheless, the other algorithms' results in Table \ref{tbl:Res_powerlaw_ne} and \ref{tbl:Res_uniform_ne} are different to those in Table \ref{tbl:Res_powerlaw} and \ref{tbl:Res_uniform}. As expected, the diversity of assignments slightly increases in terms of $H_2$, while there is a minor drop in $H_1$ values. The change is plausible since we incorporated $H_2$ into the algorithms as the fitness function instead of $H_1$.

One may observe that increasing $\frac{m}{n}$ affects the capability of the introduced algorithms in terms of $H_2$ more than it does in terms of $H_1$. This is because, in a limited feasible region, the more frequent variables are more likely to be fixed at true or false. Since those variables have a higher weight in the diversity calculation, increases in $\frac{m}{n}$ make it challenging to diversify solutions in terms of $H_2$. For instance, Table \ref{tbl:Res_powerlaw_ne} indicates that the $H_2$ values drop from $0.96$ to $0.38$ for the EDO algorithm using mutation, while the same sets of solutions result in less severe decreases in $H_1$ values (from $0.94$ to $0.5$).
 
Table \ref{tbl:Res_powerlaw_ne} also indicates that the gap between the results of the EDO algorithms' variants is more profound when $H_2$ is used as the fitness function. The statistical tests also confirm the difference in favour of the variant employing the mutation in instances where $m = \{230, 270, 290\}$. However, it is the other way around in the uniform instances; the variant that benefits from the crossover performs slightly better, although the difference is statically insignificant. The same observation we had when $H_1$ was incorporated into the algorithm as the fitness function.

\subsection{Investigation on Unsatisfiablity}
This subsection studies the correlation between the obtained diversity and the number of unsatisfiable formulas generated during the search. The introduced algorithms, as mentioned, modify the formula $\Phi$ to generate a new assignment in each iteration. Although $\Phi$ is a true formula, it is likely to make it false via modifications during the search. We consider Algorithm~\ref{alg:basic} for this purpose since the algorithm does not have any hyper-parameters affecting the results. 
\begin{figure}[t]
\centering
\includegraphics[width=0.48\columnwidth]{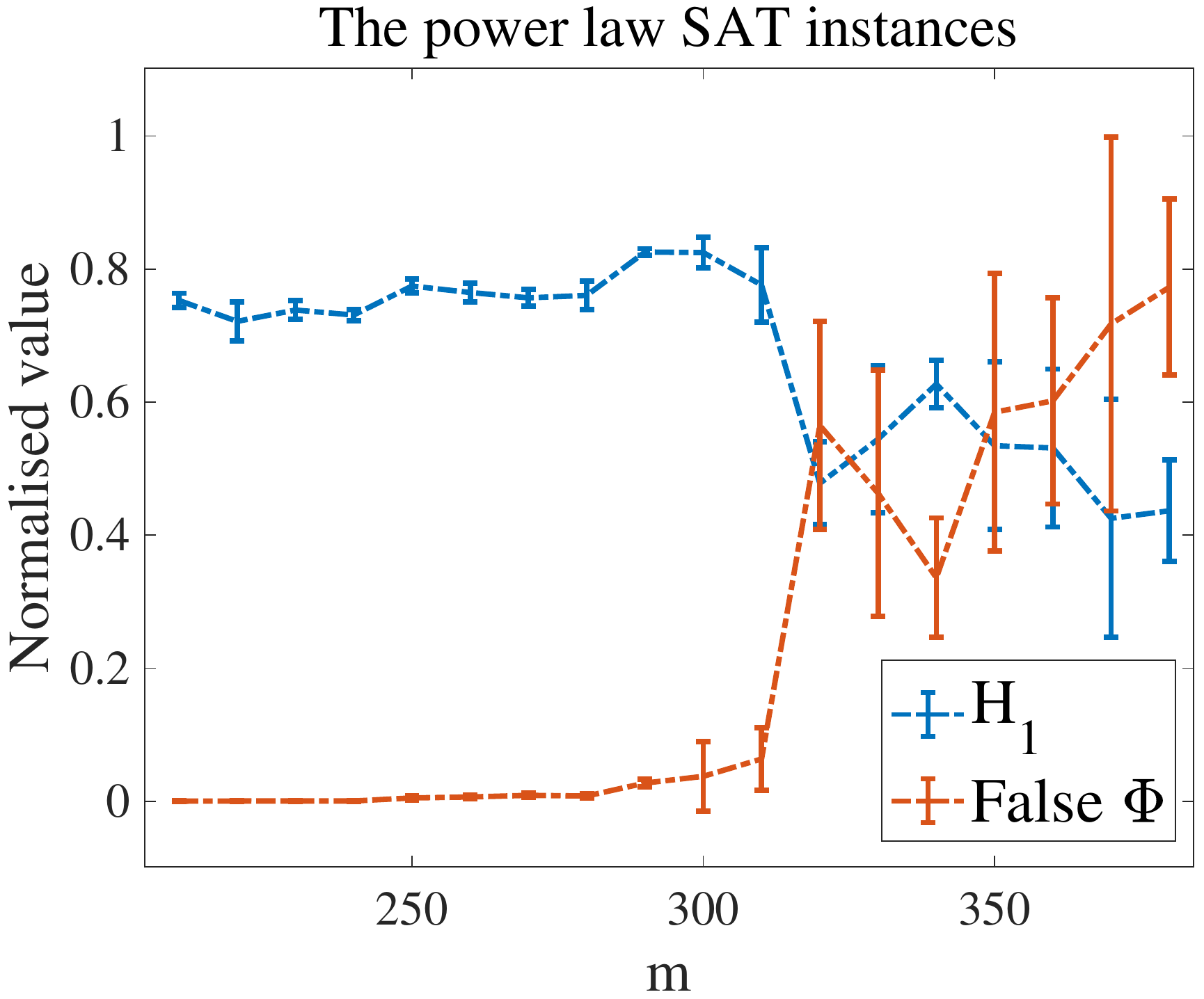}
 \hskip4pt
\includegraphics[width=0.48\columnwidth]{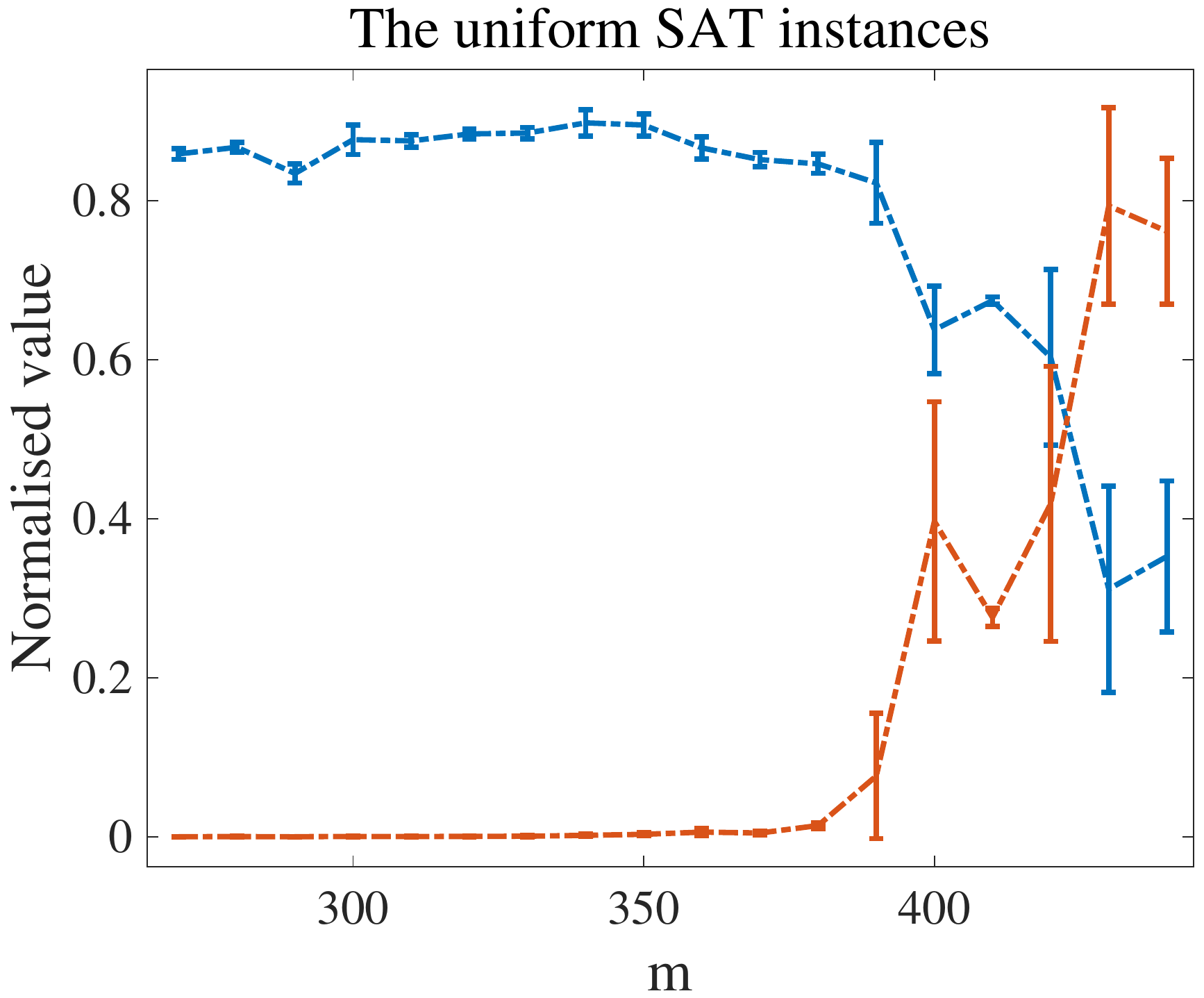}

\includegraphics[width=0.48\columnwidth]{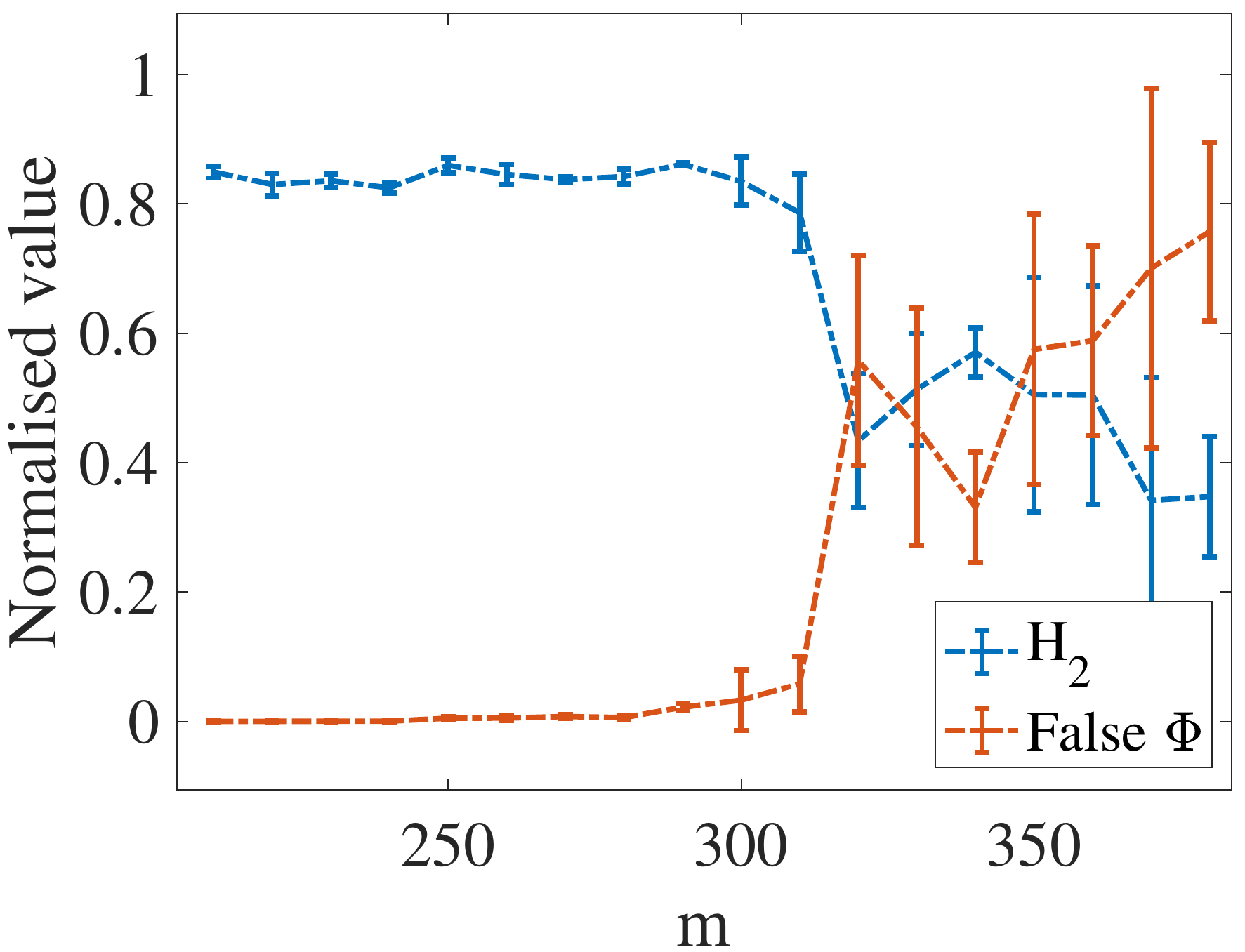}
 \hskip4pt
\includegraphics[width=0.48\columnwidth]{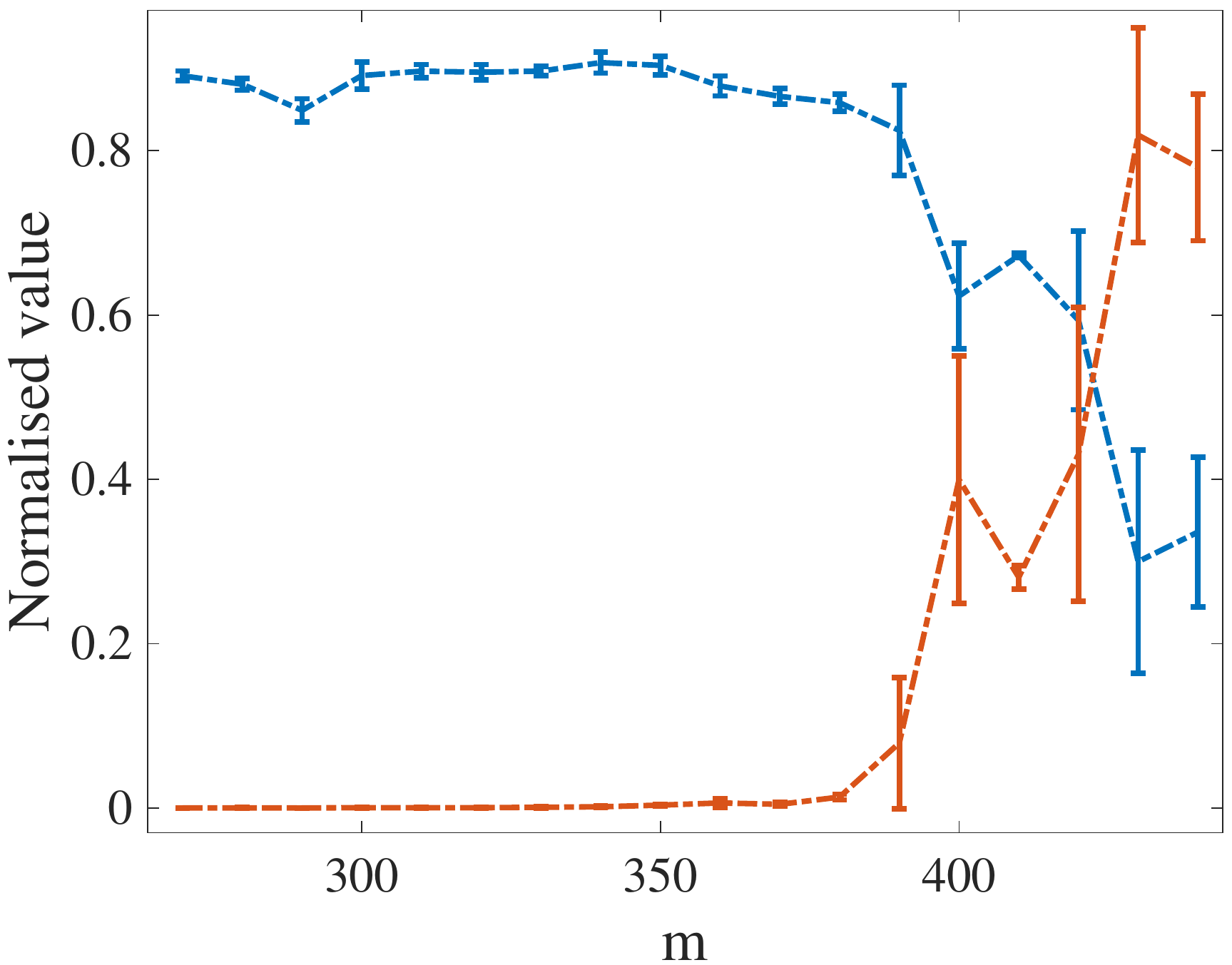}
\caption{The representative trajectories of the bit-flip algorithm's the diversity and the number of false $\Phi$. In the first row, $H_1$ serves as the fitness function, while it is $H_2$ in the second row.}
\label{fig:trend}
\end{figure}

Figure~\ref{fig:trend} depicts the trajectories of diversity and the false $\Phi$ generated by Algorithm~\ref{alg:basic}. Note that we normalise the values to plot them in a figure. As expected, the algorithm generates the minimum number of false formulas (false $\Phi$) when $\frac{m}{n}$ is low. Low values of $\frac{m}{n}$ often lead to large feasible regions and, consequently, a larger room to diversify the solutions. In such cases, the modifications of Algorithm~\ref{alg:basic} are not large enough to cause unsatisfiability for $\Phi$. If $\frac{m}{n}$ gets sufficiently large, so does the feasible region get more limited, affecting both the diversity and satisfiability rate. Although a disproportional relationship between diversity and unsatisfiability is expected, the figure interestingly depicts a symmetric behaviour. The trajectories are pretty similar for $H_1$ and $H_2$. The sole difference is the range of $H_1$ and $H_2$ in the power law instances, where $H_2$ starts and finishes at slightly higher values.    

\section{Conclusion}

This study presented evolutionary approaches to construct a diverse set of solutions in SAT using the well-known SAT solver, minisat. We first defined two measures to quantify the diversity of solutions. One, which considers and one, which dismisses the frequency of variable appearances in clauses. Then, we introduced two EAs, employing the EDO principle to construct a diverse set of SAT assignments. The EAs iteratively make modifications on a given SAT instance, then solve it with a well-known solver, minisat. Finally, we conducted a comprehensive experimental investigation to assess the algorithms' performance and study the solution space and unsatisfiability rate. The results indicate the capability of the introduced algorithm to compute highly diverse sets of SAT solutions.

For future studies, it is intriguing to study more complicated EAs, like $(\mu+\lambda)$-EAs. Although it is challenging in diversity problems to select the next generation when $\lambda$ is larger than one, an increase in $\lambda$ can potentially improve the algorithms' performance. Another possible extension is to study other related problems, such as MaxSAT.

\section*{Acknowledgements}
This work was supported by the Australian Research Council through grants DP190103894 and FT200100536.

\bibliographystyle{splncs04}
\bibliography{Main}

\end{document}